\documentclass{article}

\usepackage[preprint,nonatbib]{neurips_2026}
\usepackage[utf8]{inputenc}
\usepackage[T1]{fontenc}
\usepackage{microtype}
\usepackage{graphicx}
\usepackage{booktabs}
\usepackage{multirow}
\usepackage{array}
\usepackage{tabularx}
\usepackage{amsmath,amssymb}
\usepackage{algorithm}
\usepackage{algpseudocode}
\usepackage{placeins}
\usepackage{xcolor}
\usepackage{hyperref}

\hypersetup{
  pdftitle={Rethinking Self-Evolving Agents: Do We Still Need Prescribed Optimization Pipelines?},
  pdfauthor={Hui Xue and Fan Yang},
  colorlinks=true,
  linkcolor=blue,
  citecolor=blue,
  urlcolor=blue
}

\providecommand{\Description}[1]{}

\newcommand{\oeo}{\textnormal{\textsc{OEO}}}
\newcommand{\skillopt}{\textnormal{\textsc{SkillOpt}}}
\newcommand{\skilloptlite}{\textnormal{\textsc{SkillOpt-Lite}}}
\newcommand{\gepa}{\textnormal{\textsc{GEPA}}}
\newcommand{\best}[1]{\textbf{#1}}
\newcommand{\blocked}{\textit{blocked}}

\title{Rethinking Self-Evolving Agents:\\
Do We Still Need Prescribed Optimization Pipelines?}

\author{%
  Hui Xue\\
  Microsoft Research\\
  \texttt{xuehui@microsoft.com}\\
  \And
  Fan Yang\\
  Microsoft Research\\
  \texttt{fanyang@microsoft.com}\\
}
\date{}

\begin{document}

\maketitle

\begin{abstract}
Self-evolving agents are usually built around prescribed optimization pipelines: the framework decides how to gather evidence, revise a persistent artifact, select candidates, and stop. We ask whether this task-specific procedure remains necessary when a frontier model acts as the optimizer. We introduce Open-Ended Optimization (\oeo{}), which keeps the objective, permitted interactions, resource budget, data boundary, and evaluation fixed while allowing the optimizer to compose the improvement process online. We compare \oeo{} with two complementary prescribed approaches: \skillopt{}, a staged pipeline with bounded edits, and \gepa{}, a reflective evolutionary search. Across 14 head-to-head comparisons over 8 benchmark--target-model settings, GPT-5.5-driven \oeo{} records 12 wins, 1 tie, and 1 narrow loss of $0.21$ percentage points. It uses a median $34.3\%$ of \skillopt{}'s configured target-interaction token budget. A one-shot, zero-interaction control shows that the gains are not explained by a single prior-driven rewrite. However, delegation has a capability boundary: \skillopt{} outperforms \oeo{} with a medium optimizer, and a weak optimizer cannot operate through the unchanged \oeo{} interface. In the fully instrumented \oeo{}--\skillopt{} pair, trajectory analysis further shows that prescription changes how optimization proceeds more consistently than it changes final behavior. Together, these findings recast prescribed pipelines as capability-dependent scaffolding: essential constraints remain external, but a sufficiently capable optimizer can compose the route from measurable feedback to persistent improvement.
\end{abstract}

\section{Introduction}

Self-evolving agents promise to turn experience into persistent improvement. Instead of solving every task from scratch, an agent can revise a prompt, memory, skill, or tool-use policy that shapes its future behavior~\cite{reflexion,expel,voyager}. Most existing systems prescribe not only \emph{what} should improve, but also \emph{how} improvement should happen: which evidence to collect, how to diagnose failures, what revisions to propose, which candidates to retain, and when to stop.

This prescription appears in systems with very different mechanics. \skillopt{} uses a staged training pipeline with bounded skill edits and validation-gated updates~\cite{skillopt,skilloptrepo}. \gepa{} instead uses free-form textual mutation inside a reflective evolutionary search with Pareto-aware selection~\cite{gepa}. One is pipeline-centric and the other search-centric. Yet both make the same systems-level choice: the framework supplies the task-specific procedure that organizes learning from interaction.

This division of labor was natural when models mainly generated proposals inside an outer optimization algorithm. Frontier models challenge that premise: they can reason over traces, diagnose failures, design revisions, and adapt from feedback~\cite{opro,textgrad,gepa,adas,aflow}. With measurable outcomes, a capable optimizer may be able to organize improvement itself---choosing the evidence, hypotheses, revisions, and stopping point online.

Delegating this procedure does not remove framework governance. The framework must still fix the objective, permitted interactions, resource budget, data boundary, and evaluation. We call these fixed elements the \emph{optimization contract}. By contrast, the task-specific logic that organizes evidence, revision, selection, and stopping is the \emph{optimization meta-policy}. This distinction separates governance of the optimization problem from prescription of the route used to solve it, and leads to our central question:

\begin{quote}
\centering
\textbf{Under the same external contract, must the framework prescribe the task-specific optimization meta-policy, or can a capable optimizer compose it online?}
\end{quote}

\begin{figure*}[t]
  \centering
  \includegraphics[width=0.94\textwidth]{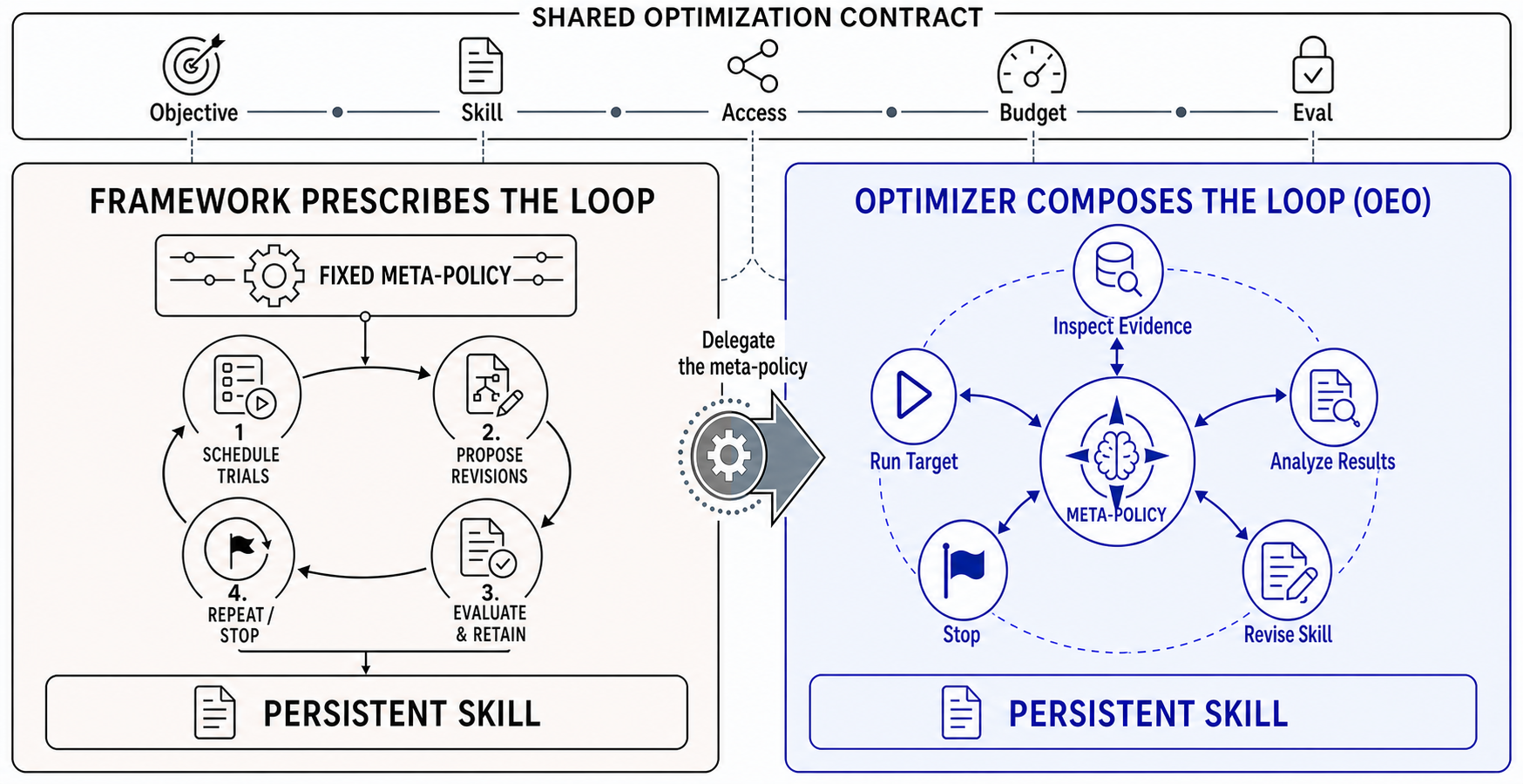}
  \caption{\textbf{Where should optimization responsibility live?} Both designs share the same external optimization contract. A prescribed pipeline fixes the task-specific improvement process in advance; \oeo{} lets a frontier optimizer compose that process online.}
  \Description{A comparison between framework-prescribed optimization and open-ended optimization. Both share an external optimization contract. In the prescribed case the framework determines the evidence, revision, selection, and stopping procedure. In OEO a frontier optimizer composes those decisions online.}
  \label{fig:concept}
\end{figure*}

We investigate this question in persistent natural-language skill optimization. A target model uses a reusable skill to solve tasks, while an optimizer improves that skill from interaction. We introduce Open-Ended Optimization (\oeo{}) as a direct test of where the meta-policy should live. \oeo{} preserves the external contract and permitted task-facing operations, but does not impose a reflection template, edit language, candidate schedule, or task-specific stopping rule. Instead, the optimizer composes the improvement process online. The comparison therefore changes the source of the improvement procedure while holding the optimization problem fixed.

With GPT-5.5 as the optimizer, \oeo{} is competitive with both representative prescribed approaches. Across 8 benchmark--target-model settings shared with \skillopt{}, it improves every initial skill and records 7 wins and 1 tie, while using a median $34.3\%$ of \skillopt{}'s configured target-interaction token budget. Across 6 confirmatory comparisons with \gepa{}, \oeo{} wins 5; \gepa{}'s only lead is $0.21$ percentage points. Because the comparators encode different edit spaces and search topologies, the result is not specific to one tested pipeline design.

We next test two explanations for this result. First, GPT-5.5 may already know how to rewrite the initial skill, making interaction unnecessary. However, a static-input-matched, zero-interaction rewrite remains below \oeo{} in all 4 tested settings and regresses on both LiveMath targets. Second, delegation may work for any optimizer. Here the pattern reverses: with the protocols frozen, \skillopt{} outperforms \oeo{} at medium optimizer capability on both ladder tasks, while a weak optimizer cannot produce a valid action through the unchanged \oeo{} interface. Thus, prescribed structure becomes useful scaffolding as the optimizer's ability to organize improvement declines.

Finally, we ask what prescription changes when it is not needed for competitive performance. In all 8 instrumented \oeo{}--\skillopt{} settings, \oeo{} makes broader maximum edits and revises more of its earlier changes, yet the selected skills often solve overlapping examples. Prescription shapes the path through skill space more consistently than the final evaluated behavior.

Taken together, these findings suggest a capability-adaptive division of labor. When feedback is measurable and the optimizer is sufficiently capable, a prescribed optimization pipeline is not a prerequisite for effective self-evolution. Instead, it becomes an optional inductive bias or a scaffold for weaker optimizers. The framework should continue to own objectives, permissions, budgets, evaluation, and governance, while delegating the task-specific route to improvement when the model can carry it. Our contributions are:

\begin{itemize}
  \item We separate the externally governed optimization contract from the task-specific optimization meta-policy, and instantiate prescribed optimization with two complementary representatives: \skillopt{} and \gepa{}.
  \item We show that a frontier optimizer can achieve competitive self-improvement without inheriting either prescribed pipeline; a matched one-shot control further shows that a single static rewrite does not reproduce the result.
  \item We identify the capability boundary of delegation and show that prescription changes committed optimization paths more consistently than final item-level behavior.
\end{itemize}

\section{Allocating Optimization Responsibility}
\label{sec:formulation}

To test this question, we vary where the task-specific meta-policy lives while keeping the optimization problem fixed. The framework always defines the problem and enforces its boundaries; only the route from evidence to a persistent update changes.

\subsection{What is fixed, and what is delegated?}

A skill-optimization instance has two model roles: the \emph{target model} $M_{\mathrm{tgt}}$ executes tasks using the current skill, while the \emph{optimizer model} $M_{\mathrm{opt}}$ analyzes interaction evidence and improves that skill. The roles may be served by the same model identity. An external optimization contract $\mathcal{C}$ fixes the objective, initial skill $s_0$, target model, permitted selection data $D_{\mathrm{sel}}$, permitted task-facing operation set $\mathcal{O}$, resource budget $B$, frozen evaluator, and sealed final split $D_{\mathrm{final}}$---but not the route from evidence to revision.

Each procedure $q$ exposes the contract's permitted operations $\mathcal{O}$ through a procedure-specific action interface $\mathcal{I}_q$, which determines its action representation, state organization, and restrictions. Given state $z_t=(s_t,H_t,K_t,b_t)$---the current skill, observations, committed checkpoints, and remaining resources---the \emph{optimization meta-policy} selects
\begin{equation}
  a_t=\mu^{(q)}(z_t;M_{\mathrm{opt}},\mathcal{C},\mathcal{I}_q),
  \qquad a_t\in\mathcal{A}_q(\mathcal{O}),
\end{equation}
where $\mathcal{A}_q(\mathcal{O})$ is the procedure-specific representation of the permitted operations. The meta-policy governs how the persistent skill is improved, not how the target model solves one task. We therefore compare complete allocations of optimization responsibility, not prompt wording alone.

\paragraph{Framework-prescribed optimization.}
A prescribed procedure fixes its action interface and substantial parts of the meta-policy outside the optimizer, including stage order, evidence routing, edit form, candidate retention, update cadence, or stopping. The optimizer supplies content within this scaffold; the framework determines how those proposals become persistent updates.

\paragraph{Open-Ended Optimization (\oeo{}).}
\oeo{} preserves the contract and a generic contract-enforcing action interface but lets $M_{\mathrm{opt}}$ compose the task-specific meta-policy online. It may gather evidence, request a targeted rollout, revise or reorganize the skill, select an earlier checkpoint, or stop. Open-ended therefore means online procedural composition, not absence of constraints.

\subsection{Two representative forms of procedural prescription}
\label{sec:prescribed-representatives}

We select two prescribed procedures with deliberately different search biases. \skillopt{} represents a \emph{staged, bounded skill-training pipeline}; \gepa{} represents a \emph{reflective evolutionary search with free-form textual mutation}. They do not share an edit language or search topology. What they share is the property under study: the framework fixes the task-specific optimization procedure before optimization begins.

\begin{table}[!t]
\centering
\caption{\textbf{Control allocation in the frontier comparison.} The contract is shared; the task-specific improvement process is controlled differently.}
\label{tab:delegation}
\footnotesize
\setlength{\tabcolsep}{5pt}
\begin{tabularx}{\textwidth}{@{}>{\raggedright\arraybackslash}p{0.19\textwidth}>{\raggedright\arraybackslash}X>{\raggedright\arraybackslash}X>{\raggedright\arraybackslash}X@{}}
\toprule
Control dimension & \skillopt{} & \gepa{} & \oeo{} \\
\midrule
External contract & Shared objective, initial skill, target, permitted data and operations, budget, selection-data boundary, and sealed final split & Same & Same \\
Evidence scheduling & Fixed rollout and reflection batches & Parent and minibatch selection & Optimizer-directed \\
Diagnosis & Success/failure decomposition & Trace-conditioned reflection & Optimizer-directed \\
Revision space & Bounded patch operators & Free-form textual mutation & Optimizer-selected edit or rewrite \\
Candidate retention & Strict improvement gate; rejected-edit memory & Pareto-aware pool update & Optimizer selects among committed states \\
Schedule and stopping & Fixed stages and epochs & Budgeted evolutionary loop & Adaptive within the budget \\
Returned skill & Validation-selected best & Validation-selected candidate & Optimizer-selected committed state \\
\bottomrule
\end{tabularx}
\end{table}

\paragraph{SkillOpt: staged and bounded skill training.}
\skillopt{} fixes rollout and reflection batches, separates success and failure diagnosis, restricts revisions to bounded patch operators, and commits only validation-improving candidates~\cite{skillopt,skilloptrepo}. It therefore represents pipeline-centric prescription: the framework determines evidence routing, edit form, acceptance, and update cadence.

\paragraph{GEPA: reflective evolutionary search.}
\gepa{} permits free-form, trace-conditioned textual mutation, but places each revision inside a framework-defined Pareto evolutionary loop~\cite{gepa}. It therefore represents search-centric prescription: the model controls mutation content, while the framework controls parent selection, evaluation cadence, population retention, and termination.

Together, the comparators span a constrained patch-based pipeline and a flexible population-based search. All three methods begin from the same artifact, receive the same data and permitted task-facing operations, and consolidate experience into a persistent skill. Thus, our comparison changes how improvement is organized, not whether experience persists; alignment details appear in Appendix~\ref{app:protocol}.

\subsection{The Open-Ended Optimization protocol}

\oeo{} is a contract-enforced agent loop; Algorithm~\ref{alg:oeo} in Appendix~\ref{app:protocol} gives the complete pseudocode. At each turn, the optimizer observes the current skill, accumulated evidence, committed checkpoints, and remaining resources, then chooses one permitted operation. The runner validates and executes that operation, records its result and cost, and prevents access to the sealed final split. A revision is committed immediately for subsequent rollouts; the optimizer may later select any committed checkpoint as its output.

Thus, \oeo{} delegates procedural composition while the runner continues to enforce the external contract.

\section{Empirical Findings}
\label{sec:results}

We first establish the frontier comparison, then test whether the result can be explained by a one-shot prior and whether it persists across optimizer capability.

\subsection{A frontier optimizer can improve skills without prescribed pipelines}
\label{sec:main-result}

\paragraph{Headline finding.}
Without inheriting either prescribed procedure, \oeo{} records 7 wins and 1 tie against \skillopt{} across 8 benchmark--target-model settings, and 5 wins against \gepa{} in 6 confirmatory settings; \gepa{}'s only lead is $0.21$ percentage points. Meanwhile, \oeo{} uses a median $34.3\%$ of the configured \skillopt{} target-interaction token budget. A frontier optimizer can therefore organize competitive skill improvement without either tested prescribed pipeline.

\begin{figure}[H]
\centering
\includegraphics[width=0.98\textwidth]{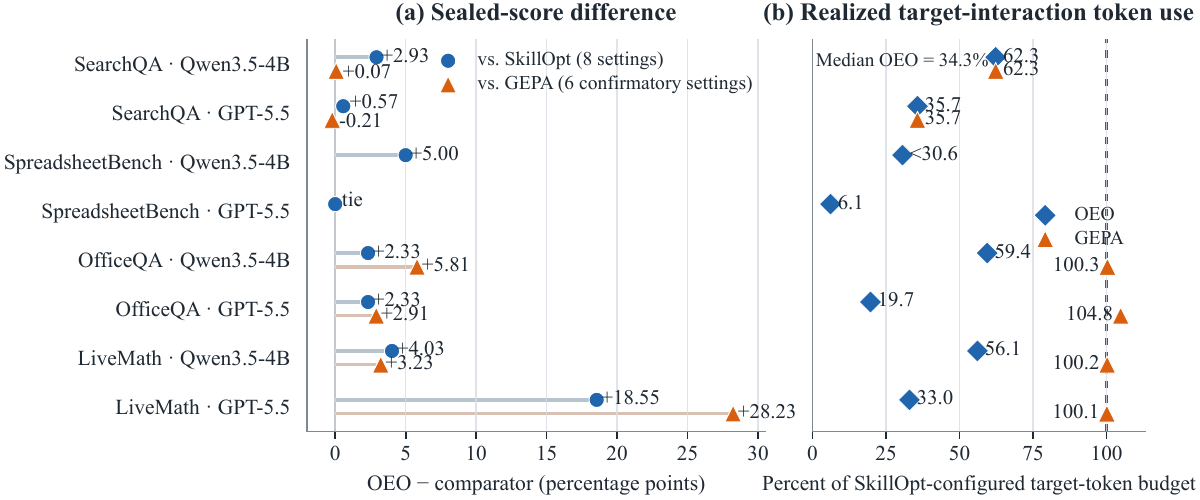}
\caption{\textbf{Relative performance and target-interaction use.} Left: sealed-score differences between \oeo{} and all 8 \skillopt{} cells and the 6 confirmatory \gepa{} cells; deltas use unrounded pass rates. Right: realized target-interaction token use as a fraction of the configured \skillopt{} reference budget; the dashed line marks $100\%$. SpreadsheetBench \gepa{} diagnostics are omitted.}
\Description{Two panels summarize the frontier comparison. The left panel plots OEO score differences against SkillOpt and GEPA for each benchmark and target model. The right panel plots OEO and GEPA target-interaction token use as a percentage of the configured SkillOpt reference budget.}
\label{fig:headline-results}
\end{figure}

\paragraph{Evaluation scope.}
The comparison spans the official SearchQA and OfficeQA evaluation splits, SpreadsheetBench, and LiveMathematicianBench (abbreviated LiveMath)~\cite{searchqa,spreadsheetbench,officeqapro,officeqarepo,livemath}, with Qwen3.5-4B and GPT-5.5 as target models~\cite{qwen35,gpt55}. Within each setting, every method starts from the same skill and operates under the same optimization contract. In Table~\ref{tab:exact-scores}, \emph{Target} denotes the model that executes benchmark tasks; GPT-5.5 supplies all optimizer-side calls for every method and every row.\footnote{All GPT-5.5 calls use the GitHub Copilot Responses API.} Table~\ref{tab:exact-scores} establishes absolute self-improvement, while Figure~\ref{fig:headline-results} summarizes relative performance margins and target-interaction token use. Full protocol and accounting details appear in Appendix~\ref{app:protocol}.

\begin{table}[t]
\centering
\caption{\textbf{Sealed-split scores under the frontier optimizer, rounded to four decimals.} Here $n$ is the number of final-evaluation items. All methods start from the same initial skill, and bold marks the best optimized score in each row. The 6 SearchQA, OfficeQA, and LiveMath \gepa{} cells are confirmatory. $\dagger$ marks post-hoc OEO-matched SpreadsheetBench checkpoints, excluded from confirmatory counts.}
\label{tab:exact-scores}
\small
\setlength{\tabcolsep}{7pt}
\begin{tabular}{llrrrrr}
\toprule
Benchmark & Target & $n$ & Initial & \skillopt{} & \gepa{} & \oeo{} \\
\midrule
SearchQA & Qwen3.5-4B & 1,400 & 0.6693 & 0.7221 & 0.7507 & \best{0.7514} \\
SearchQA & GPT-5.5 & 1,400 & 0.7857 & 0.8657 & \best{0.8736} & 0.8714 \\
\midrule
SpreadsheetBench & Qwen3.5-4B & 280 & 0.2107 & 0.1964 & 0.2214$^{\dagger}$ & \best{0.2464} \\
SpreadsheetBench & GPT-5.5 & 280 & 0.3964 & \best{0.7607} & 0.4964$^{\dagger}$ & \best{0.7607} \\
\midrule
OfficeQA & Qwen3.5-4B & 172 & 0.2616 & 0.2849 & 0.2500 & \best{0.3081} \\
OfficeQA & GPT-5.5 & 172 & 0.5872 & 0.7035 & 0.6977 & \best{0.7267} \\
\midrule
LiveMath & Qwen3.5-4B & 124 & 0.3065 & 0.6129 & 0.6210 & \best{0.6532} \\
LiveMath & GPT-5.5 & 124 & 0.3871 & 0.5000 & 0.4032 & \best{0.6855} \\
\bottomrule
\end{tabular}
\end{table}

\paragraph{Performance across complementary comparators.}
The margins are heterogeneous but the pattern spans both forms of prescription. Against \skillopt{}, \oeo{} ranges from a tie on GPT-5.5-target SpreadsheetBench to a lead of $18.55$ percentage points on GPT-5.5-target LiveMath. Against \gepa{}, SearchQA differs by at most $0.21$ percentage points, while the largest lead is $28.23$ percentage points on GPT-5.5-target LiveMath. \gepa{} itself exceeds \skillopt{} in 3 of the 6 confirmatory cells, so the extension is not explained by a uniformly weak second comparator. From the common initial skill, \oeo{} improves all 8 settings, \skillopt{} improves 7, and \gepa{} improves 5 of its 6 confirmatory settings.

\paragraph{Selective target interaction.}
\oeo{} uses a median $34.3\%$ of the configured target-interaction token budget and stays below it in every setting. On SearchQA, \oeo{} and \gepa{} use the same target-interaction token fraction and reach nearly identical scores. Across OfficeQA and LiveMath, \gepa{} reaches roughly $100\%$ in all 4 runs, whereas \oeo{} uses $19.7\%$--$59.4\%$ and scores higher. On GPT-5.5-target LiveMath, \oeo{} uses $33.0\%$ of the reference allocation while producing the largest performance gap; on GPT-5.5-target SpreadsheetBench, it uses $6.1\%$ and ties \skillopt{} at $0.7607$. Thus, the result is not explained by spending more target-interaction tokens.

This efficiency is specific to target interaction. \oeo{} slightly exceeds the configured \skillopt{} optimizer-token budget in 3 settings and reaches $98.9\%$ in a fourth. The resource pattern is therefore selective target interaction, not uniform token reduction. Overall, GPT-5.5 builds a competitive route to skill improvement online without either tested task-specific procedure.

\subsection{A single zero-interaction rewrite does not explain the gains}
\label{sec:one-shot-control}

Table~\ref{tab:one-shot} asks whether GPT-5.5 can obtain the same gains from its prior alone. Static knowledge helps on SearchQA: the one-shot rewrite improves the initial skill by $2.07$ and $5.00$ percentage points. It still remains $6.14$ and $3.57$ percentage points below \oeo{}. On LiveMath, the same control lowers both initial scores, while \oeo{} improves them by $34.68$ and $29.84$ percentage points.

\begin{table}[H]
\centering
\caption{\textbf{Static-input-matched one-shot control.} Each cell uses one GPT-5.5 rewrite and zero target-model optimization rollouts. Bold marks the highest score in each row.}
\label{tab:one-shot}
\small
\setlength{\tabcolsep}{3.2pt}
\begin{tabular}{llrrr}
\toprule
Benchmark & Target & Initial & One-shot rewrite & \oeo{} \\
\midrule
SearchQA & Qwen3.5-4B & 0.6693 & 0.6900 & \best{0.7514} \\
SearchQA & GPT-5.5 & 0.7857 & 0.8357 & \best{0.8714} \\
LiveMath & Qwen3.5-4B & 0.3065 & 0.2742 & \best{0.6532} \\
LiveMath & GPT-5.5 & 0.3871 & 0.3468 & \best{0.6855} \\
\bottomrule
\end{tabular}
\end{table}

A single static frontier-model rewrite therefore does not reproduce \oeo{} in any tested cell. Frontier-model prior helps---especially on SearchQA---but the interactive loop supplies additional gains that a single static rewrite does not reproduce.\footnote{The GPT-5.5-target LiveMath one-shot evaluation contains 2 item-level timeouts counted as failures. Even treating both as correct would raise its score only to $0.3629$, leaving \oeo{} ahead by $32.26$ percentage points.}

\subsection{Delegation is sensitive to optimizer capability}
\label{sec:capability}

With the protocols frozen, Table~\ref{tab:capability} shows a clear crossover on both ladder tasks. At frontier capability, \oeo{} leads \skillopt{} on both; at medium capability, \skillopt{} leads by $9.68$ percentage points on LiveMath and $3.50$ percentage points on SearchQA. Under the unchanged \oeo{} interface, the weak optimizer emits non-executable actions and no rollout occurs; the \skillopt{} runner completes both runs by supplying the control sequence externally. Executability alone does not guarantee useful improvement: weak-optimizer \skillopt{} scores $0.6579$ on SearchQA, below the shared initial-skill score of $0.6693$.

\begin{table}[H]
\centering
\caption{\textbf{Frozen-protocol optimizer-capability ladder.} The target is Qwen3.5-4B. Bold marks the higher score when both methods complete; \emph{blocked} means that no valid executable action was produced and therefore no target rollout occurred.}
\label{tab:capability}
\small
\setlength{\tabcolsep}{2.5pt}
\begin{tabular}{llccc}
\toprule
Benchmark & Method & \shortstack{Weak\\Qwen3.5-4B} & \shortstack{Medium\\Qwen3.5-27B} & \shortstack{Frontier\\GPT-5.5} \\
\midrule
\multirow{2}{*}{LiveMath}
& \oeo{} & \blocked{} & 0.4919 & \best{0.6532} \\
& \skillopt{} & 0.5565 & \best{0.5887} & 0.6129 \\
\midrule
\multirow{2}{*}{SearchQA}
& \oeo{} & \blocked{} & 0.6807 & \best{0.7514} \\
& \skillopt{} & 0.6579 & \best{0.7157} & 0.7221 \\
\bottomrule
\end{tabular}
\end{table}

Thus, \oeo{} is not an optimizer-independent wrapper. Moving from the medium to the frontier optimizer raises \oeo{} by $16.13$ percentage points on LiveMath and $7.07$ percentage points on SearchQA, compared with $2.42$ and $0.64$ percentage points for \skillopt{}. The larger sensitivity of \oeo{} shows that online meta-policy composition depends directly on optimizer capability. The same optimization responsibility cannot simply be assigned unchanged across capability levels.

\section{What Changes Under a Prescribed Procedure?}
\label{sec:path-function}

The frontier comparison shows that a prescribed pipeline is not necessary for competitive performance. We next ask what prescription changes inside the optimization process by analyzing the fully instrumented \oeo{}--\skillopt{} pair.

\paragraph{Headline finding.}
Prescription changes the optimization path more consistently than it changes the selected skill's evaluated behavior. Across all 8 instrumented settings, \oeo{} makes a larger maximum edit, touches a broader set of sections, and exhibits greater revision churn than \skillopt{}. Yet pass/fail agreement is at least $0.78$ in 7 of 8 settings, and correct-set Jaccard exceeds $0.70$ in 5. In this pair, \skillopt{}'s prescribed pipeline is associated with a narrower, less revision-heavy observed path, but it does not uniquely determine the evaluated behavior reached.

\paragraph{Diagnostics.}
We reconstruct every byte-distinct committed skill state and compare the selected final skills at the item level. Path measures capture the largest normalized edit, the broadest Markdown-section change, and revision churn; outcome measures capture pass/fail agreement and correct- and gain-set Jaccard. Equivalent checkpoint instrumentation is unavailable for \gepa{}, so this diagnostic analysis uses the fully logged \oeo{}--\skillopt{} pair. Appendix~\ref{app:path-metrics} gives the complete definitions.

\subsection{Different committed paths}

\begin{figure}[!t]
\centering
\includegraphics[width=\linewidth]{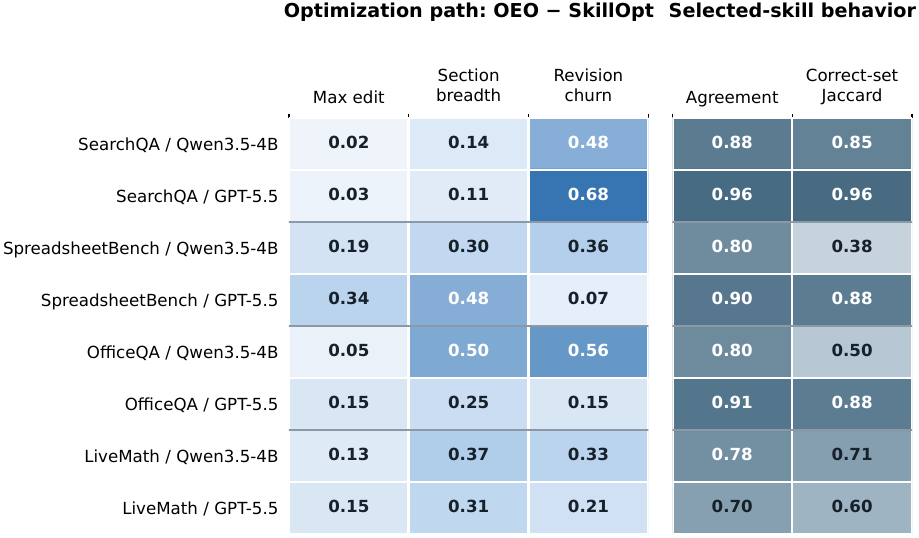}
\caption{\textbf{Path divergence does not imply divergent evaluated behavior.} The first 3 columns are \oeo{} minus \skillopt{} committed-path metrics and share a common color scale; all differences are positive. The final columns show pass/fail agreement and correct-set Jaccard ($J_C$) between selected skills.}
\Description{An annotated matrix with 8 benchmark and target-model rows. All OEO-minus-SkillOpt path differences are positive, while selected-skill pass-fail agreement and correct-set Jaccard vary across settings.}
\label{fig:path-function}
\end{figure}

The trajectories reject a simple ``fewer, larger updates'' story. \oeo{} has more committed updates in 5 settings, the same number in 1, and fewer in 2. By contrast, all 3 path metrics move in one direction: \oeo{} has a larger maximum normalized token edit, larger maximum section breadth, and higher revision churn in all 8 paired runs (Figure~\ref{fig:path-function}). Relative to \skillopt{}, \oeo{} makes broader maximum moves and later rewrites more of the text introduced along its path, consistent with the difference between online composition and staged, gated editing. Thus, in this pair, prescription shapes the route through artifact space rather than simply reducing the number of updates.

\subsection{Different committed paths can yield overlapping item-level behavior}

Path metrics summarize chronological committed states, whereas final behavior is measured on each procedure's selected skill. Table~\ref{tab:functional} reports their item-level overlap. Pass/fail agreement ranges from $0.7016$ to $0.9629$, while correct-set Jaccard ranges from $0.3778$ to $0.9581$. On SearchQA with GPT-5.5, agreement is $0.9629$ and correct-set Jaccard is $0.9581$ despite different histories. SpreadsheetBench with GPT-5.5 is especially informative: the methods tie at $0.7607$, with $0.9000$ agreement and $0.8767$ correct-set Jaccard. Distinct text paths do not merely reach the same leaderboard number; the selected skills succeed on largely the same evaluated instances.

\begin{table}[!t]
\centering
\caption{\textbf{Final item-level correctness overlap in the instrumented \oeo{}--\skillopt{} pair.} Scores are rounded to four decimals. Agreement is binary pass/fail agreement; $J_C$ is correct-set Jaccard; $J_G$ is gain-set Jaccard relative to the shared initial skill. Higher values indicate greater overlap.}
\label{tab:functional}
\small
\setlength{\tabcolsep}{6pt}
\begin{tabular}{llccccc}
\toprule
Benchmark & Target & \oeo{} & \skillopt{} & Agreement & $J_C$ & $J_G$ \\
\midrule
SearchQA & Qwen3.5-4B & 0.7514 & 0.7221 & 0.8793 & 0.8486 & 0.5567 \\
SearchQA & GPT-5.5 & 0.8714 & 0.8657 & 0.9629 & 0.9581 & 0.7500 \\
\midrule
SpreadsheetBench & Qwen3.5-4B & 0.2464 & 0.1964 & 0.8000 & 0.3778 & 0.2653 \\
SpreadsheetBench & GPT-5.5 & 0.7607 & 0.7607 & 0.9000 & 0.8767 & 0.8136 \\
\midrule
OfficeQA & Qwen3.5-4B & 0.3081 & 0.2849 & 0.8023 & 0.5000 & 0.3000 \\
OfficeQA & GPT-5.5 & 0.7267 & 0.7035 & 0.9070 & 0.8779 & 0.7273 \\
\midrule
LiveMath & Qwen3.5-4B & 0.6532 & 0.6129 & 0.7823 & 0.7065 & 0.7500 \\
LiveMath & GPT-5.5 & 0.6855 & 0.5000 & 0.7016 & 0.5978 & 0.3617 \\
\bottomrule
\end{tabular}
\end{table}

The exceptions show that different routes can also yield different correctness patterns. LiveMath/GPT-5.5 has a large score gap, with agreement $0.7016$ and correct-set Jaccard $0.5978$. OfficeQA/Qwen3.5-4B has relatively close aggregate scores but a correct-set Jaccard of only $0.5000$, showing that the methods solve meaningfully different item sets. Similar aggregate performance therefore does not imply overlapping evaluated behavior. Conversely, large textual differences need not imply behavioral divergence: across the 8 settings, final-text distance and behavioral distance have a descriptive Spearman correlation of only $\rho=0.3095$. Optimization path, skill text, and final behavior therefore capture different properties and should be evaluated separately.

\FloatBarrier
\section{Discussion}

\paragraph{Procedural support should follow optimizer capability.}
Our results reposition prescribed procedures as capability-dependent support rather than the sole source of optimization intelligence. At frontier capability, \oeo{} is competitive with two structurally different prescribed procedures. At lower capability, however, the fixed \skillopt{} pipeline becomes advantageous, and the weak optimizer cannot operate through the unchanged \oeo{} interface. Prescription is therefore neither universally necessary nor uniformly redundant: it can scaffold optimization when the model cannot reliably compose the improvement process for itself.

\paragraph{Delegating the meta-policy does not remove framework governance.}
Delegation leaves the external contract unchanged: objectives, permissions, budgets, and evaluation boundaries remain framework-owned. Only the task-specific sequence---which evidence to inspect, which revision to try, and when to stop---moves to the optimizer. This distinction suggests a capability-adaptive design: begin with delegated composition when the optimizer can use executable feedback effectively, and introduce stronger procedural support when it cannot reliably produce or select useful updates.

\paragraph{Process and function require separate evaluation.}
Skill text, committed trajectory, and evaluated behavior are distinct objects. Different procedures can traverse substantially different paths yet reach overlapping item-level behavior, while similar aggregate scores can conceal complementary correct-item sets. Self-evolution should therefore be evaluated with both process-level diagnostics and item-level outcomes; neither a text diff nor a final benchmark score alone explains what the system learned.

\section{Related Work}

\paragraph{Persistent improvement without weight updates.}
Reflexion stores verbal self-reflections, ExpeL extracts reusable natural-language insights, and Voyager grows an executable skill library through interaction~\cite{reflexion,expel,voyager}. These systems establish that persistent external state can improve future behavior, but they optimize different artifacts and organize updates differently.

\paragraph{Language models as optimizers.}
OPRO proposes candidates from evaluated solutions, Promptbreeder evolves task and mutation prompts, and TextGrad propagates textual feedback through compound systems~\cite{opro,promptbreeder,textgrad}. ADAS uses a meta-agent to program agent designs, while AFlow searches code-represented workflows~\cite{adas,aflow}. These works demonstrate optimization-level reasoning by language models. We ask a different systems question: must the framework still prescribe the task-specific improvement procedure?

\paragraph{Prescribed agent and skill evolution.}
Bayesian-Agent maps belief states to structured skill operations~\cite{bayesianagent}. \skillopt{} prescribes staged rollout, reflection, bounded editing, and validation-gated updates; \gepa{} permits free-form mutations inside a Pareto evolutionary loop~\cite{skillopt,skilloptrepo,gepa}. \skilloptlite{} retains trajectory exploration, consensus mining, and independent validation in a lighter pipeline~\cite{skilloptlite}. SkillOS, HDSO, and SkillAudit place structure in curator learning, prospective hypothesis testing, or paired verification and rollback~\cite{skillos,hdso,skillaudit}. These methods motivate procedural prescription as a broad design axis, not a single algorithmic family.

\section{Conclusion}

Self-evolving agents commonly place the procedure for learning from experience in the framework. We separate the externally governed optimization contract from the task-specific meta-policy that organizes improvement. Across the tested settings, a frontier-driven \oeo{} improves every initial skill and remains competitive with two different prescribed procedures; a static-input-matched zero-interaction rewrite does not reproduce its gains.

Prescription nevertheless remains useful. In the frozen-protocol ladder, \skillopt{} leads on both tasks at medium capability, while its runner remains executable when the weak optimizer is blocked under the unchanged \oeo{} interface. In the instrumented frontier pair, \oeo{} and \skillopt{} traverse different committed paths while often reaching overlapping evaluated behavior. The resulting principle is capability-adaptive responsibility: keep the contract external, and treat task-specific prescription as a design choice rather than a default when progress is measurable. Delegate online composition only to an optimizer that can carry it effectively.

\appendix

\section{Detailed Experimental Protocol}
\label{app:protocol}

\subsection{Benchmarks, splits, and evaluation}

Optimization data supply task evidence, and the permitted selection split---usually named \texttt{valid\_seen}---may guide candidate or checkpoint selection. This selection-data boundary specifies which evidence can guide the returned skill. The sealed \texttt{valid\_unseen} or test split is used once after selection and supplies the primary pass-rate metric under each benchmark's frozen evaluator. Each reported numerical score corresponds to one completed optimization run under the recorded configuration; blocked cells contain no score. The sealed evaluations contain SearchQA ($n=1{,}400$), SpreadsheetBench ($n=280$), OfficeQA ($n=172$), and LiveMath ($n=124$) examples. Within a benchmark--target pair, all compared methods use the same dataset snapshot and split, target endpoint, initial skill, and evaluator.

\subsection{Frontier comparison and resource accounting}

The primary experiment uses GPT-5.5 as the optimizer-side model for all three procedures. The complete \oeo{}--\skillopt{} comparison contains 8 benchmark--target-model settings. The aligned \gepa{} condition evolves only the common persistent skill, starts from the byte-identical artifact, disables optional merging, and returns one validation-selected candidate. The 6 confirmatory runs cover SearchQA, OfficeQA, and LiveMath with both targets. SpreadsheetBench reports a post-hoc candidate from the checkpoint matched to \oeo{} target-interaction token use; because this is not \gepa{}'s native full-budget selection, those 2 cells are descriptive rather than confirmatory.

We record optimizer tokens separately from target-model tokens spent during optimization and checkpoint-selection rollouts. \skillopt{}'s configured allocations define the common reference budget, and \oeo{} and \gepa{} report realized use relative to it. Model calls are atomic, so a call that begins within the budget can produce a small overshoot when it completes. The tables therefore compare sealed performance and realized token use separately rather than claiming exact realized-cost matching.

\subsection{One-shot and capability controls}

The zero-interaction rewrite receives the byte-identical initial skill and the same static step-0 information as \oeo{}: benchmark and target identity, task and interface specification, objective, budget fields, evaluator and output requirements, and tool-availability description. Tool access is then disabled. It cannot inspect examples, execute the target, observe trajectories or scores, or revise more than once. Each cell contains one GPT-5.5 completion and zero target-model optimization rollouts.

For the capability ladder, Qwen3.5-4B is the target on SearchQA and LiveMath, while the optimizer varies across Qwen3.5-4B, Qwen3.5-27B, and GPT-5.5. Prompts, target endpoint, data, seed, action interface, and method protocol remain fixed within each method. Weak and medium \oeo{} caps match the realized frontier-\oeo{} token budgets; \skillopt{} retains its configured pipeline budget. Invalid actions are not repaired and count as operational failures.

\begin{algorithm}[t]
\caption{Contract-enforced Open-Ended Optimization}
\label{alg:oeo}
\footnotesize
\begin{algorithmic}[1]
\Require contract $\mathcal{C}$ containing $s_0$, budget $B$, and sealed split $D_{\mathrm{final}}$; optimizer $M_{\mathrm{opt}}$
\State $s\gets s_0$; history $H\gets\emptyset$; checkpoints $K\gets[s_0]$; selected skill $\hat{s}\gets\bot$
\While{the budget admits another optimizer step}
  \State $a_t\gets M_{\mathrm{opt}}(s,H,K,\mathrm{remaining}(B);\mathcal{C})$
  \State validate $a_t$ against the permitted interface and remaining resources
  \If{$a_t=\textsc{Revise}(s')$}
    \State $s\gets s'$; append $s$ to $K$; record the commit in $H$
  \ElsIf{$a_t=\textsc{Select}(j)$}
    \State $\hat{s}\gets K[j]$; record the selection in $H$
  \ElsIf{$a_t=\textsc{Stop}$}
    \State \textbf{break}
  \Else
    \State execute \textsc{Inspect} or \textsc{RunTarget}; append the observation to $H$
  \EndIf
  \State update optimizer and target-model resource counters
\EndWhile
\State \Return $\hat{s}$ if selected, otherwise $s$; evaluate it once on sealed $D_{\mathrm{final}}$
\end{algorithmic}
\end{algorithm}

\subsection{Trajectory and item-level diagnostics}
\label{app:path-metrics}

For each run, we reconstruct the initial skill $s_0$ and every byte-distinct committed state in chronological order, ending at $s_T$. Let $\operatorname{Lev}(s,s')$ denote token-level Levenshtein distance and $|s|$ the number of tokens. For adjacent committed states, normalized edit is $\operatorname{Lev}(s_{t-1},s_t)/\max(|s_{t-1}|,|s_t|,1)$. A Markdown section is identified by its full heading path; section breadth is the fraction of paths in the union of two adjacent states whose bodies change or whose paths are added or removed. We report the maximum normalized edit and maximum section breadth along each run. Revision churn is
\begin{equation}
\mathrm{Churn}=1-\frac{\operatorname{Lev}(s_0,s_T)}{\max\!\left(\sum_{t=1}^{T}\operatorname{Lev}(s_{t-1},s_t),1\right)},
\end{equation}
which measures how much stepwise editing does not survive in the terminal state. Path metrics use chronological committed states, whereas behavioral and final-text metrics use the selected returned skills $s_O^\star$ and $s_S^\star$. Their normalized final-text distance is $d_{\mathrm{text}}=\operatorname{Lev}(s_O^\star,s_S^\star)/\max(|s_O^\star|,|s_S^\star|,1)$. Pass/fail agreement is the fraction of sealed examples on which the two selected skills have identical binary correctness, and behavioral distance is $d_{\mathrm{behavior}}=1-\text{agreement}$. If $C_0$, $C_O$, and $C_S$ are the sealed correct-item sets of the initial, \oeo{}, and \skillopt{} skills, define gain sets $G_O=C_O\setminus C_0$ and $G_S=C_S\setminus C_0$. We report correct-set Jaccard $J_C=|C_O\cap C_S|/|C_O\cup C_S|$ and gain-set Jaccard $J_G=|G_O\cap G_S|/|G_O\cup G_S|$, with Jaccard defined as one when both sets are empty. Rejected and uncommitted proposals are excluded.

\end{document}